\documentclass[10pt,twocolumn,letterpaper]{article}

\usepackage{wacv}
\usepackage{float}
\usepackage{multirow}
\usepackage{tabularx}
\usepackage{booktabs} 
\usepackage[T1]{fontenc} 
\usepackage{listings}
\usepackage{xcolor}
\usepackage{afterpage}
\usepackage{stfloats} % Allows [b] tracking for double-column floats
\usepackage{caption}  % For proper caption handling
\usepackage[table]{xcolor}
\definecolor{wacvblue}{rgb}{0.21,0.49,0.74}
\usepackage[pagebackref,breaklinks,colorlinks,allcolors=wacvblue]{hyperref}

\def\wacvPaperID{*****} % *** Enter the WACV Paper ID here
\def\confName{WACV}
\def\confYear{2027}

\title{MIRCaps: A Large-Scale Mixed-Domain Dataset with Image-Level and Region-Level Captions for Fine-Grained Vision-Language Learning}

\author{Arlindo Luciano Tulumba Roberto\\
Changwon National University\\
Changwon City, South Korea\\
{\tt\small 20257901@cs.cwnu.ac.kr}
\and
Hyungjoon Kim\\
Changwon National University\\
Changwon City, South Korea\\
{\tt\small hyungjoon@changwon.ac.kr}
}

\begin{document}
\maketitle
\begin{abstract}
Despite recent progress in Vision-Language Models (VLMs), mixed-domain image-caption datasets for both general-purpose and CCTV-based video surveillance systems remain limited. To address this gap, we introduce a large-scale multimodal dataset comprising 141,364 images, 981,947 image-level captions, 1,742,264 region-level captions, and 1,391,779 bounding box annotations. Each image is associated with an average of seven image-level captions describing different aspects of the overall scene, as well as seven region-level captions for each annotated bounding box. These complementary caption types are designed to help VLMs learn fine-grained visual attributes, including object categories, estimated sizes, colors, actions, states, and surrounding environmental context. We demonstrate the effectiveness of the dataset on two important downstream tasks: image captioning and object detection. Experimental results show that lightweight VLMs, including SmolVLM-256M-Instruct, BLIP, BLIP2, and Qwen2.5-VL-3B-Instruct, can be effectively fine-tuned using our dataset. Our dataset and code are publicly available at \url{https://zenodo.org/records/20418601}.
\end{abstract}
    
\section{Introduction}
\label{sec:intro}

Recent advancements in Vision-Language Models (VLMs) have revolutionized the video analytics field. Architectures such as BLIP-2, LLaVA, and Flamingo effectively bridge visual perception with natural language reasoning, enabling video analytics systems to generate dense scene descriptions, execute visual question answering (VQA), perform text-to-video retrieval, and conduct complex scene-level reasoning. For instance, lightweight perception modules can be coupled with VLM-based semantic reasoning to mitigate hallucinations and improve real-time CCTV surveillance and retail event understanding \cite{chen2026smarteyes}. However, optimization of VLMs requires more than scaling data volume; it fundamentally hinges on the granularity and precision of the underlying text-image pairings. Prior works demonstrate that explicit modeling of object attributes, spatial topologies, actions, object states, and environmental context is critical for robust cross-modal alignment and high-fidelity caption generation \cite{li2024object,kaduri2025whats}. Unfortunately, standard benchmarks like MS COCO \cite{chen2015coco} heavily underrepresent these nuances, typically providing short, generic annotations. A classic example \textit{“A large bus sitting next to a very tall building”}, fails to articulate vital, application-specific variables such as vehicle color, operational status (e.g., parked vs. moving, damaged, or crowded), or fine-grained human activities (e.g., boarding passengers), which are crucial for downstream video understanding.

To address the limited availability of mixed-domain multimodal datasets for fine-grained vision-language learning, we make three primary contributions:

\begin{enumerate}
    \item \textbf{Large-scale mixed-domain multimodal dataset:} We introduce MIRCaps, a large-scale mixed-domain multimodal dataset containing 141,364 images, 981,947 image-level captions, 1,742,264 region-level captions, and 1,391,779 bounding box annotations spanning both general-purpose and CCTV-domain images.

    \item \textbf{Object-centric captions:} We provide multiple object-centric region captions for each annotated object, enabling fine-grained supervision of visual attributes such as object category, estimated size, color, action, state, and surrounding environmental context.

    \item \textbf{Comprehensive evaluation of lightweight Vision-Language Models:} We conduct a comprehensive evaluation of lightweight Vision-Language Models, including SmolVLM-256M-Instruct, BLIP, BLIP2, and Qwen2.5-VL-3B-Instruct, using both human evaluation and standard automatic metrics, including BLEU-4, CIDEr-D, METEOR, BLIPScore, and BERTScore. Although not the primary focus of this work, we additionally evaluate object detection models, including YOLOv12s and RT-DETR-L, to assess the quality and utility of the provided bounding box annotations.    
\end{enumerate}

\begin{figure}[H]
  \centering
   \includegraphics[width=0.8\linewidth]{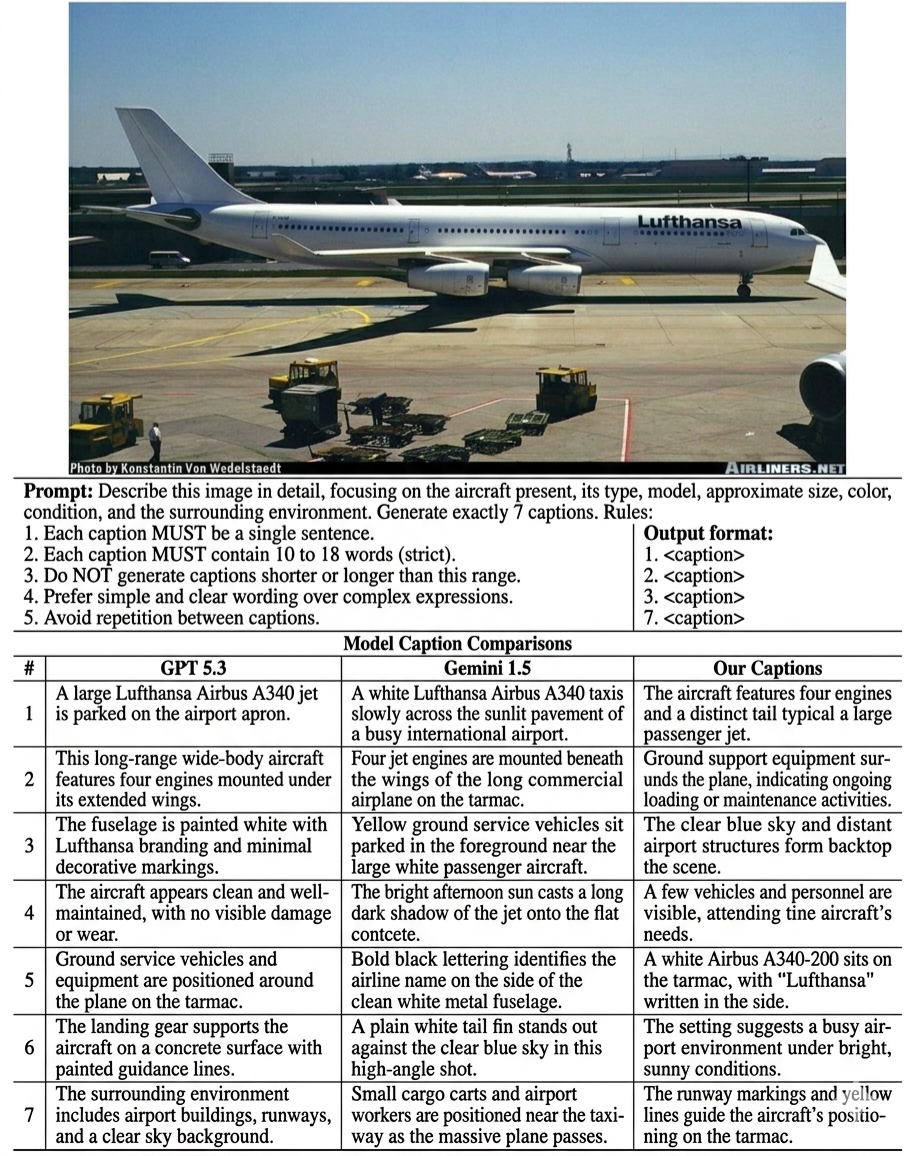}

   \caption{Comparison between Image-level captions generated by SOTA Multimodal Models and Our captions. We provide a clearer view of this sample in Appendix~\ref{sec:appendix_qualitative_analysis} (Figure~\ref{fig:comparison_sota}).}
   \label{fig:onecol}
\end{figure}

Despite using a significantly smaller model (InternVL3.5-14B) for caption generation, our dataset demonstrates strong semantic quality by consistently capturing key attributes such as the primary object, type, estimated size, color, and scene, as well as fine-grained details like exact aircraft models (e.g., \textit{A340-200}), which larger models often omit, highlighting the overall quality and effectiveness of the dataset.

%-------------------------------------------------------------------------
\section{Methodology}

\subsection{Dataset Collection}

\subsubsection{Definition of Object Categories}

One of the objectives of this work is to evaluate how effectively lightweight vision–language models such as SmolVLM-256M-Instruct, BLIP, BLIP2, and Qwen2.5-VL-3B-Instruct can generate descriptions of both different aspects of image content and object-specific attributes, conditioned on object categories detected by YOLOv12 or RF-DETR-v2. So, for the object detection task, we define 15 categories representing common objects observable in CCTV footage, aligned with our project requirements: \textit{Person (0), Head (1), Bird (2), Fire (3), Smoke (4), Helmet (5), Vest (6), Car (7), Bus (8), Truck (9), Vehicle (10), Bicycle (11), Motorcycle (12), Aircraft (13)}, and \textit{Watercraft (14)}. The numbers in parentheses denote category IDs. While most categories are self-explanatory, Head refers specifically to a human head, as defined in the CrowdHuman dataset \cite{shao2018crowdhuman}, and Vehicle denotes any motorized vehicle other than a car (e.g., buses and trucks), consistent with the OD-VIRAT-Tiny dataset \cite{ullah2011odvirat}.

\subsubsection{Image and Bounding Box Annotation Sources and Selection Criteria}
To construct the dataset, we collected images and bounding box annotations from several publicly available object detection datasets, each containing one or more of the defined categories. These include FGVC-Aircraft dataset \cite{maji2013fgvc}, DFire dataset \cite{venancio2022automatic}, DFS Fire and Smoke dataset \cite{wu2022fire}, CrowdHuman dataset \cite{shao2018crowdhuman}, OD-VIRAT-Tiny dataset \cite{ullah2011odvirat}, Forest Fire dataset \cite{shamta2023forest}, CQUniversity Fire and Smoke dataset \cite{partheepan2025annotated}, CALTECH CUB-200-2011 Bird dataset \cite{wah2011cub}, Vehicle Orientation dataset \cite{kumar2022traffic}, MS COCO (2014) dataset \cite{lin2014coco}, McShips dataset [11] \cite{zheng2020mcships}, Personal Protective Equipment (PPE) dataset \cite{huang2025ppe}, and HardHat-Vest Dataset \cite{zahit2023hardhat}. Apart from category coverage and dataset size, other criteria used for selecting these datasets included the presence of single or multiple objects within a scene, as well as the need to ensure diversity and variability across the final image-caption data. Refer to Appendix~\ref{sec:appendix_dataset_sources}
(Table~\ref{tab:dataset_sources}) for additional details on the original dataset sources, associated licenses, and download URLs.

\subsection{Dataset Preprocessing}
\textbf{File Filtering}. 
To standardize the dataset, files with inconsistent labeling across images, missing image or label files, and duplicate entries were removed prior to processing.

\textbf{Conversion of Bounding Box Coordinates}. 
Datasets whose bounding box annotations are not originally provided in YOLO format are converted to a unified representation. For example, the OD-VIRAT-Tiny dataset stores annotations in COCO format [x, y, w, h], where pair x, y denotes the top-left corner, and all annotations are contained within a single JSON file. Each annotation entry includes an image identifier, category identifier, and a bounding box specified as [x, y, w, h]. We transform these annotations into the YOLO format [xcenter, ycenter, w, h], where all coordinates are normalized to [0,1] with respect to the image width and height. The converted annotations are stored as individual text (\texttt{.txt}) files per image, with each line formatted as: \texttt{<class\_ID>} \texttt{<x\_center>} \texttt{<y\_center>} \texttt{<w>} \texttt{<h>}.

\textbf{Category Name and ID Mapping}. 
Additionally, a unified category name and ID mapping scheme was adopted to ensure consistency across the dataset. Any irrelevant categories in the source datasets were ignored and therefore not mapped to the final dataset. Appendix~\ref{sec:appendix_dataset_preprocessing} (Table~\ref{tab:category_id_mapping}) provides additional details on this topic.

\textbf{File Renaming}. 
Filenames were largely preserved, with only minor adjustments for consistency and compatibility. In CrowdHuman, commas were replaced with underscores (e.g., \textit{273271,1a0d6000b9e1f5b7.jpg} $\rightarrow$ \textit{273271\_1a0d6000b9e1f5b7.jpg}). In OD-VIRAT-Tiny, filenames were prefixed for dataset identification (e.g., \textit{0.jpg} $\rightarrow$ \textit{od\_virat\_tiny\_0.jpg}). In DFS Fire and Smoke, redundant suffixes were removed (e.g., \textit{large\_-1-\_jpg.rf.f3d78ad3c2ef9677dc1a12393d51378b.jpg} $\rightarrow$ \textit{large\_1.jpg}).

\subsection{Caption Generation and Cleansing}
\label{sec:caption_generation_cleansing}
In this section, we describe the caption generation and cleansing procedures used to construct the dataset.

\subsubsection{Caption Generation}
We employed InternVL3.5-14B model to generate image captions for all images. To enrich the semantic quality of the generated descriptions, additional metadata from the original datasets was incorporated into the prompting process. Specifically, bird species information from the CALTECH-200-2011 Birds dataset, aircraft manufacturer and model types from the FGVC-Aircraft dataset, and ship categories (e.g., civilian or warship) from the McShips dataset were injected into the prompts. Furthermore, prompts were dynamically constructed based on object categories derived from bounding box annotations, enabling the model to produce context-aware and category-relevant captions. 

%%%%%%%%%%%%%%%%%%%%%%%%%%%%%%%%%%%%%%%%%%%%%%%%%%%%%%%
\begin{figure}[h]
\centering
\includegraphics[width=\columnwidth]{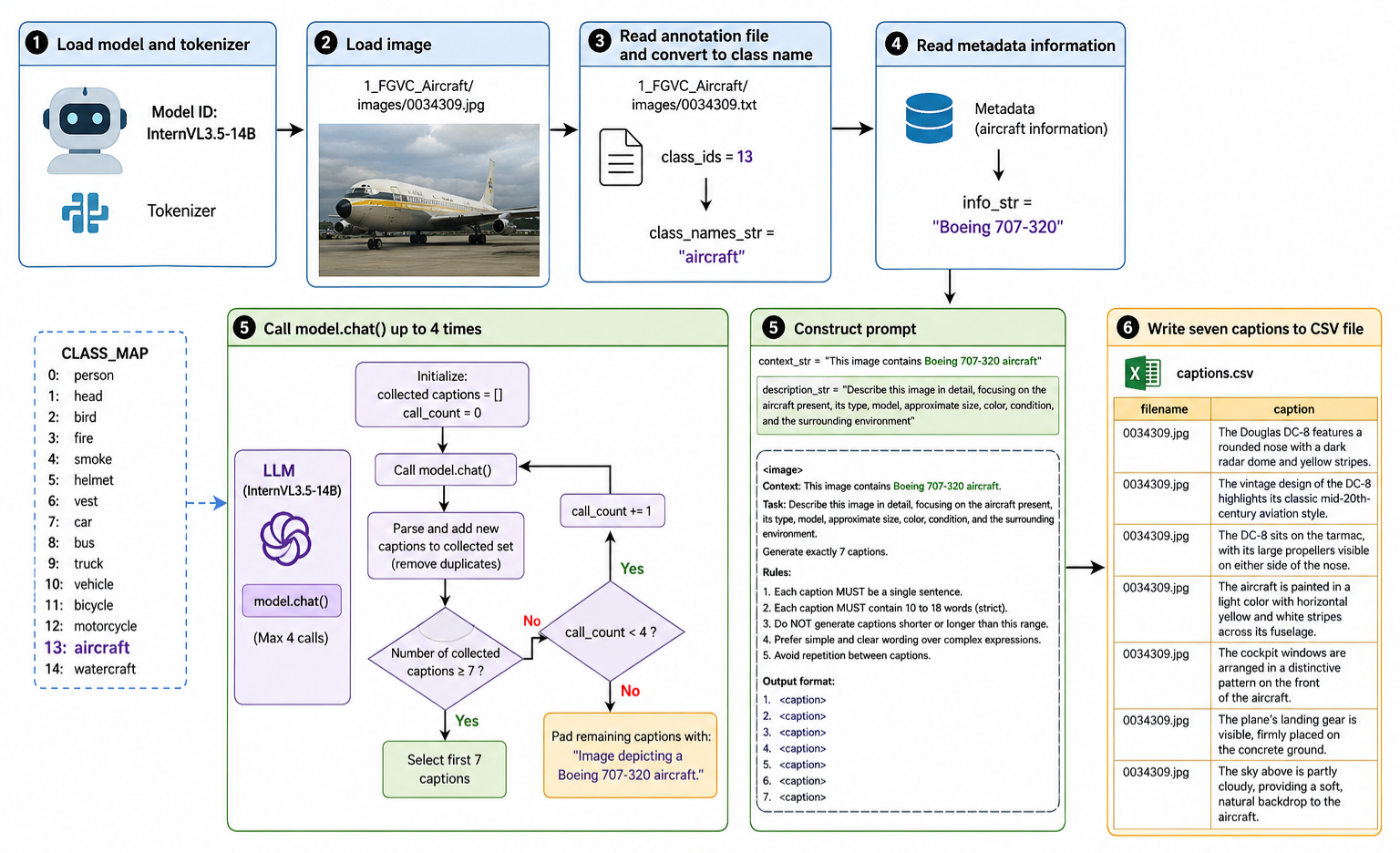}
\caption{Caption Generation Pipeline.}
\label{fig:caption_generation_strategy_1}
\end{figure}
%%%%%%%%%%%%%%%%%%%%%%%%%%%%%%%%%%%%%%%%%%%%%%%%%%%%%%%
Figure~\ref{fig:caption_generation_strategy_1} illustrates the caption generation strategy applied for the generation of both image and region-level captions. An image from the FGVC-Aircraft dataset is used for illustration purposes. We provide a higher-resolution view of this sample in Appendix~\ref{sec:appendix_caption_generation} (Figure~\ref{fig:caption_generation_strategy_2}).

\subsubsection{Caption Cleansing}
To clean the caption dataset, we removed exact duplicate captions for the same image or region, as well as extra spaces, inconsistent punctuation, and capitalization errors.

\subsection{Dataset Organization and Statistics}
\label{sec:dataset_organization_statistics}
\noindent\textbf{Dataset Organization}. 
Our dataset is organized into 13 directories, each corresponding to a specific dataset listed in Table~\ref{tab:dataset_statistics}. Each dataset directory contains an \textit{images} directory with the images and a \textit{labels} directory with bounding box annotations in YOLO format. It also includes a CSV file for image-level captions and another for region-level captions, along with an additional CSV file providing auxiliary metadata. This metadata file is available only for three datasets: FGVC-Aircraft, McShips, and CALTECH CUB-200-2011 Bird.

\begin{figure}[h]
\centering
\includegraphics[width=\columnwidth]{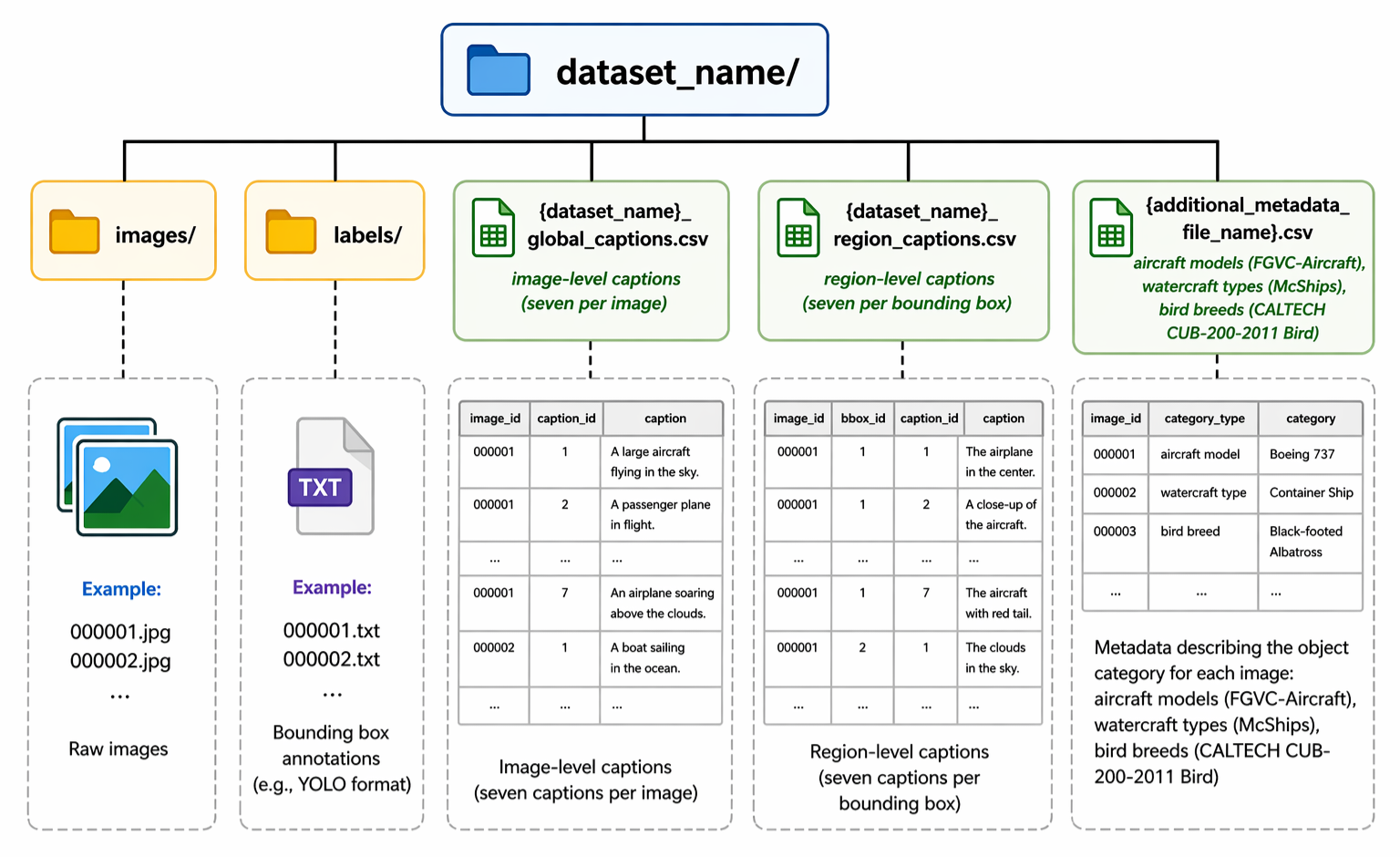}
\caption{Basic Dataset Structure.}
\label{fig:dataset_structure}
\end{figure}

For instance, the \textit{FGVC-Aircraft} dataset is organized under \textit{MIRCaps/datasets/1\_FGVC\_Aircraft}. It includes an \textit{images} directory with 10,000 images and a \textit{labels} directory with 10,000 corresponding \textit{.txt} annotation files. Additionally, it provides \textit{1\_fgvc\_aircraft\_global\_captions.csv} (70,000 image-level captions) and \textit{1\_fgvc\_aircraft\_region\_captions.csv} (70,000 region-level captions). An extra metadata file, \textit{1\_fgvc\_aircraft\_models.csv}, contains aircraft manufacturer and model information for 6,668 images. For example, the entry \textit{1025794.jpg, "Boeing 707-320"} indicates that the aircraft in image \textit{1025794.jpg} is a \textit{Boeing 707-320}. Other datasets follow the same basic directory structure.

\textbf{Dataset Statistics}. 
As illustrated in Table~\ref{tab:dataset_statistics}, the \textit{DFire} dataset contains the highest volume of both images and bounding box annotations (21,527) and global captions (149,323), while the \textit{Forest Fire} dataset holds the absolute lowest across all three of these metrics, with 1,690 images and bounding box annotations and 11,830 global captions. When shifting focus to localized annotations, the \textit{CrowdHuman} dataset features the highest density of region captions at 248,768, whereas the \textit{McShips} dataset registers the lowest at 55,153. Lastly, for supplementary data, the \textit{McShips} dataset dominates with the highest volume of metadata info at 78,171 files, while ten out of the thirteen sub-datasets have no metadata files.

\begin{table*}[t]
\centering
\scriptsize
\caption{Dataset Statistics}
\label{tab:dataset_statistics}
% Scales the table exactly to the width of the text block
\resizebox{\textwidth}{!}{%
\begin{tabular}{p{3.3cm} p{2.6cm} r r r r r}
\hline
\textbf{Dataset Name} & \textbf{Folder Name} & \textbf{\# Images} & \textbf{\# Annotations} & \textbf{\# Global Captions} & \textbf{\# Region Captions} & \textbf{\# Metadata Info} \\
\hline

FGVC-Aircraft dataset \cite{maji2013fgvc} &
\texttt{1\_FGVC\_Aircraft} &
10,000 & 10,000 & 70,000 & 69,723 & 6,668 \\

DFire dataset \cite{venancio2022automatic} &
\texttt{2\_DFire} &
21,527 & 21,527 & 149,323 & 185,753 & 0 \\

DFS Fire and Smoke dataset \cite{wu2022fire} &
\texttt{3\_DFS\_Fire\_Smoke} &
8,735 & 8,735 &  61,135 & 131,925 & 0 \\

CrowdHuman dataset \cite{shao2018crowdhuman} &
\texttt{4\_CrowdHuman} &
20,396 & 20,396 & 135,549  & 248,768 & 0 \\

OD-VIRAT-Tiny dataset \cite{ullah2011odvirat} &
\texttt{5\_OD\_VIRAT\_Tiny} &
19,860 & 19,860 & 139,018 & 159,211 & 0 \\

Forest Fire dataset \cite{shamta2023forest} &
\texttt{6\_Forest\_Fire} &
1,690 & 1,690 & 11,830 & 27,411 & 0 \\

CQUniversity Fire and Smoke dataset \cite{partheepan2025annotated} &
\texttt{7\_CQ\_Fire\_Smoke} &
4,954 & 4,954 & 34,678 & 77,473 & 0 \\

CALTECH CUB-200-2011 Bird dataset \cite{wah2011cub} &
\texttt{8\_CALTECH\_Bird} &
11,787 & 11,787 & 82,509 & 82,463  & 11,787 \\

Vehicle Orientation dataset \cite{kumar2022traffic} &
\texttt{9\_VEHICLE\_Orientation} &
12,940 & 12,940 & 90,580 & 238,834 & 0 \\

MS COCO (2014) dataset \cite{lin2014coco} &
\texttt{10\_MS\_COCO\_2014} &
12,583 & 12,583 & 88,081 & 199,088 & 0 \\

McShips dataset \cite{zheng2020mcships} &
\texttt{11\_MC\_Ships} &
7,879 & 7,879 & 55,153 & 55,153 & 78,171 \\

Personal Protective Equipment (PPE) dataset \cite{huang2025ppe} &
\texttt{12\_PPE} &
2,286 & 2,286 & 16,002 & 26,604 & 0 \\

HardHat-Vest Dataset \cite{zahit2023hardhat} &
\texttt{13\_HardHat\_Vest} &
6,727 & 6,727 & 47,089 & 216,840 & 0 \\

\hline

\textbf{Total} &
\textbf{13 Folders} &
\textbf{141,364} &
\textbf{141,364} &
\textbf{980,947} &
\textbf{1,719,246} &
\textbf{96,626} \\

\hline
\end{tabular}%
}
\end{table*}

\begin{table*}[t]
\centering
\scriptsize
\caption{Comparison between our dataset and well-known benchmark datasets}
\label{tab:dataset_comparison}
\begin{tabularx}{\textwidth}{>{\raggedright\arraybackslash}p{2.8cm} r r c c c c >{\raggedright\arraybackslash}X}
\hline
\textbf{Dataset} &
\textbf{\# Images} &
\textbf{\# Captions} &
\textbf{Avg. Caps/Img} &
\textbf{Avg. Cap. Length} &
\textbf{Cap. Category} &
\textbf{Annot. Method} &
\textbf{Img Domain} \\
\hline

LAION-400M \cite{schuhmann2021laion} &
400,000,000 &
400,000,000 &
1 &
$\sim$10--15 &
Image-level &
Automatic &
Gen. purpose \\

Conceptual Captions \cite{sharma2018conceptual} &
3,300,000 &
3,300,000 &
1 &
9.7 &
Image-level &
Automatic &
Gen. purpose \\

MS COCO Captions \cite{chen2015coco} &
330,000 &
1,650,000 &
5 &
10.5 &
Image-level &
Manual &
Gen. purpose \\

SBU Captions \cite{ordonez2011im2text} &
1,000,000 &
1,000,000 &
1 &
12.0 &
Image-level &
Automatic &
Gen. purpose \\

\textbf{Ours} &
\textbf{141,364} &
\textbf{981,947} &
\textbf{6.99} &
\textbf{12.8} &
\textbf{Image-level} &
\textbf{Automatic} &
\textbf{CCTV, Gen. purpose} \\

Flickr30k \cite{plummer2015flickr30k} &
31,783 &
158,915 &
5 &
12--13 &
Image-level &
Manual &
Gen. purpose \\

\textbf{Ours} &
\textbf{141,364} &
\textbf{1,742,264} &
\textbf{6.98} &
\textbf{12.32} &
\textbf{Region-level} &
\textbf{Automatic} &
\textbf{CCTV/Gen. purpose} \\

Visual Genome \cite{krishna2017visual} &
108,000 &
5,400,000 &
50 &
5.0 &
Region-level &
Manual &
Gen. purpose \\

NoCaps \cite{agrawal2019nocaps} &
15,100 &
166,100 &
11 &
11.0 &
Region-level &
Manual &
Gen. purpose \\

TextCaps \cite{sidorov2020textcaps} &
28,000 &
145,000 &
5 &
12.4 &
Region-level &
Manual &
Gen. purpose (OCR) \\

GroundCap \cite{oliveira2025groundcap} &
52,016 &
52,016 &
1 &
128.0 &
Region-level &
Mixed &
Movie \\

\hline
\end{tabularx}
\end{table*}

\textbf{Comparison with Benchmark Datasets}. 
We provide a comprehensive comparative analysis of our dataset against established image-captioning benchmarks (Table~\ref{tab:dataset_comparison}), evaluating key dimensions such as image and caption scale, average number of captions per image, average sentence length, types of captions, and annotation methodology. The listed benchmarks are organized by the total number of captions. Regarding image-level annotations, our dataset provides 981,947 image-level captions, significantly exceeding Flickr30k (158,915), while LAION-400M provides the largest overall volume (400 million). For region-level annotations, despite Visual Genome having the largest amount (5.4 million), our dataset contains 1,742,264 region-level captions, which substantially outperforms NoCaps (166,100), demonstrating a much higher annotation density.

\subsection{Data Quality Assessment}

\subsubsection{Image Quality Assessment}
In this section, we evaluate the visual quality of our image dataset using both no-reference image quality assessment (NR-IQA) metrics, specifically NIQE~\cite{mittal2013blind}, BRISQUE~\cite{mittal2012noreference}, and PIQE~\cite{venkatanath2015blind}, and low-level visual attribute analyses, such as brightness, contrast, and sharpness. We also evaluate fundamental statistical characteristics of the dataset, including resolution distributions, aspect ratios, color (RGB) channels, and data integrity metrics such as the corrupted image ratio. Additional details are provided in Appendix~\ref{sec:image_quality_assessment} (Tables~\ref{tab:nriqa}, \ref{tab:visual_attributes}, and~\ref{tab:image_dataset_characteristics}).

\subsubsection{Bounding Box Annotation Quality Assessment}
\textbf{Bounding Box Visual Analysis:}
Before training the object detection models (YOLOv12s and RT-DETR-L), we conducted a visual inspection of the bounding box annotations across all source datasets. The inspection indicated accurate object labeling and no critical mismatches between annotated objects and their corresponding bounding boxes. 

\begin{figure}[h]
\centering
\includegraphics[width=\columnwidth]{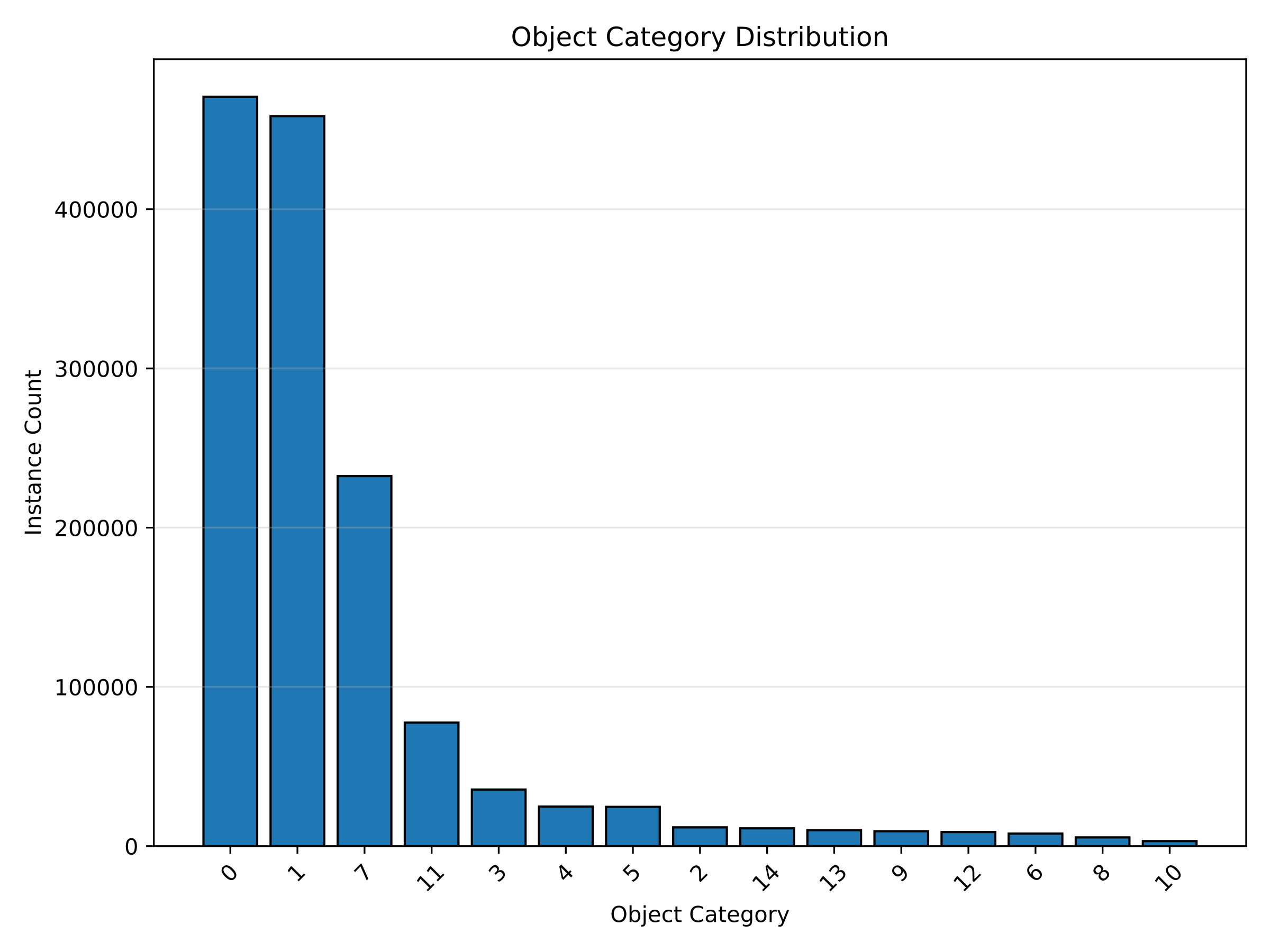}
\caption{Object Category Distribution.}
\label{fig:dataset_categorydistrib}
\end{figure}

\textbf{Category Distribution Analysis:}
The dataset contains 1,391,779 bounding box instances across 15 object categories (IDs 0–14). As shown in Figure~\ref{fig:dataset_categorydistrib}, it exhibits a characteristic long-tail distribution reflecting real-world urban frequency, with a heavy emphasis on human-centric categories. This highly imbalanced distribution is dominated by Person (0: 470,648) and Head (1: 458,434), while other categories such as Car (7: 232,410) and Bicycle (11: 77,550) are moderately represented, and rare classes like Bus (8: 5,475) and Vehicle (10: 3,158) occupy the tail of the distribution with substantially fewer instances.

\subsubsection{Caption Quality Assessment}
\label{sec:human_evaluation_caption_quality}
For caption quality assessment, we employ two evaluation methods. First, we use CLIPScore \cite{hessel2021clipscore} to measure image–caption alignment across the entire dataset, ensuring that generated captions achieve sufficient semantic consistency for downstream training and applications. Second, we conduct human evaluation to verify that captions satisfy our dataset requirements, including accurate descriptions of key attributes such as primary objects, approximate size, color, actions, states, and surrounding context. 

\textbf{CLIPScore-based Caption Quality Assessment:}
Table~\ref{tab:clipscore_category} summarizes the semantic alignment quality of the proposed dataset across the two caption categories using the CLIPScore metric computed with Equation~\ref{eq:clipscore_image}. The dataset achieves an overall average CLIPScore of 28.56 over 1,131,257 image-caption pairs. Region-level captions attain a slightly higher CLIPScore of 28.62 on 141,709 region-caption pairs, outperforming the image-level captions, which achieve 28.50 across 981,947 image-caption pairs. This result indicates marginally stronger semantic correspondence for localized region descriptions compared to full-image captions.
%%%%%%%%%%%%%%%%%%%%%%%%%%%%%%%%%%%%%%%%%%%%%%%%%%%%
\begin{equation}
  CS_{\mathrm{image}} = \frac{1}{N_{\mathrm{image}}} \cdot \sum_{i=1}^{N_{\mathrm{image}}} \mathrm{CLIPScore}(I_i, C_i)
  \label{eq:clipscore_image}
\end{equation}
%%%%%%%%%%%%%%%%%%%%%%%%%%%%%%%%%%%%%%%%%%%%%%%%%%%%

\noindent
where:
\begin{itemize}
    \item $CS_{\mathrm{image}}$ denotes the final CLIPScore,
    \item $I_i$ denotes the image,
    \item $C_i$ denotes the corresponding image-caption pair,
    \item $N_{\text{image}}$ denotes the total number of image-caption pairs.
\end{itemize}

\begin{table}[h]
  \centering
  \scriptsize
  \begin{tabular}{@{}lcc@{}}
    \toprule
    \textbf{Caption Category} & \textbf{Avg. CLIPScore} & \textbf{\# Image-Caption Pairs} \\
    \midrule
    Image-level & 28.50 & 981,947 \\
    Region-level & \textbf{28.62} & 141,709 \\
    \midrule
    \textbf{Overall} & \textbf{28.56} & \textbf{1,131,257} \\
    \bottomrule
  \end{tabular}
  \caption{CLIPScore by Caption Category.}
  \label{tab:clipscore_category}
\end{table}

\textbf{Human Evaluation of Caption Quality:}
To validate semantic accuracy and caption fidelity, we randomly selected 142 images and 142 cropped images from each of the 13 source datasets in MIRCaps to ensure both dataset diversity and an equal number of samples across all datasets. This process resulted in a human evaluation subset containing 25,844 image-caption pairs, including 12,922 image-level captions and 12,922 region-level captions. Three independent annotators evaluated each image-caption pair across eight dimensions: \textit{Hallucination Check}, \textit{Object Presence Consistency}, \textit{Size}, \textit{Color}, \textit{Action}, \textit{State}, \textit{Scene Context}, and \textit{Additional Information}. The purpose of each evaluation category can be inferred directly from its corresponding question presented in Table~\ref{tab:human_evaluation_caption_quality}.

%%%%%%%%%%%%%%%%%%%%%%%%%%%%%%%%%%%%%%%%%%%%%%%%%%%%%%%%%%%%%%%%
\begin{table*}[t]
  \centering
  \small
  \begin{tabularx}{\textwidth}{@{}l X l@{}}
    \toprule
    \textbf{Category} & \textbf{Evaluation Question} & \textbf{Response} \\
    \midrule

    Hallucination Check &
    \textbf{Q1:} Are all objects, actions, or attributes mentioned in the captions visually present in the image? &
    YES / NO \\
    \addlinespace

    Object Presence Consistency &
    \textbf{Q2:} Can every object named in the captions be verified in the image? &
    YES / NO \\
    \addlinespace

    Size &
    \textbf{Q3:} Is there at least one caption that accurately describes the size of a primary object? &
    YES / NO \\
    \addlinespace

    Color &
    \textbf{Q4:} Is there at least one caption that accurately describes the color of a primary object? &
    YES / NO \\
    \addlinespace

    Action &
    \textbf{Q5:} Is there at least one caption that accurately describes the action of a primary object? &
    YES / NO \\
    \addlinespace

    State &
    \textbf{Q6:} Is there at least one caption that accurately describes the state or condition of a primary object? &
    YES / NO \\
    \addlinespace

    Scene Context &
    \textbf{Q7:} Is there at least one caption that accurately describes the scene or environment in the image? &
    YES / NO \\
    \addlinespace

    Additional Information &
    \textbf{Q8:} Is there at least one caption describing additional details such as airplane model, bird breed, ship type (civilian ship, warship), etc., beyond the primary object's basic features? &
    YES / NO \\

    \bottomrule
  \end{tabularx}
  \caption{Human Evaluation Criteria for Caption Quality.}
  \label{tab:human_evaluation_caption_quality}
\end{table*}
%%%%%%%%%%%%%%%%%%%%%%%%%%%%%%%%%%%%%%%%%%%%%%%%%%%%%%%%%%%%%%%%
Table~\ref{tab:per_question_overall_performance} shows the human evaluation results for caption quality assessment. YES responses were mapped to 1 and NO responses to 0. Final scores were computed as the percentage of YES responses across all annotators and evaluated image-caption pairs. For each evaluation question, the corresponding score was computed using Equation~\ref{eq:question_score}, while the overall average score was calculated using Equation~\ref{eq:overall_average}.

\begin{equation}
\text{Question Score} =
\left(
\frac{\sum \text{YES responses}}
{N_{\text{samples}} \times N_{\text{annotators}}}
\right)
\times 100
\label{eq:question_score}
\end{equation}

\begin{equation}
\text{Overall Average} =
\left(
\frac{\sum \text{All YES responses}}
{\sum \text{All evaluations}}
\right)
\times 100
\label{eq:overall_average}
\end{equation}

\begin{table}[t]
\centering
\caption{Per Question Overall Performance. Here, IC and RC denote image-level and region-level captions. Results in this table are derived from the individual human evaluation results for image-level and region-level captions, provided in Appendix~\ref{sec:appendix_human_evaluation_caption_quality} (Tables~\ref{tab:image_level_caption_performance} and \ref{tab:region_level_caption_performance}). Strongest performance exceeding 90\% are shown in \textbf{bold}.}

\footnotesize
\renewcommand{\arraystretch}{1.15}
\setlength{\tabcolsep}{3pt}

\begin{tabular*}{\columnwidth}{@{\extracolsep{\fill}}lcc}
\hline
\textbf{Human Evaluation Questions} & \textbf{IC (\%)} & \textbf{RC (\%)} \\
\hline

Q1 (Hallucination Check) & \textbf{97.33} & 74.32 \\
Q2 (Object Presence Consist.) &  \textbf{92.24} & 71.52 \\
Q3 (Size) & 22.01 & 35.26 \\
Q4 (Color) &  \textbf{93.86} & 90.68 \\
Q5 (Action) & 61.21 & 37.45 \\
Q6 (State) & 70.91 & 57.37 \\
Q7 (Scene Context) &  \textbf{99.13} & 58.16 \\
Q8 (Additional Information) &  \textbf{95.02} & 32.56 \\
\hline

\textbf{Overall Avg.} & \textbf{78.96} & \textbf{57.17} \\
\hline
\label{tab:per_question_overall_performance}
\end{tabular*}
\end{table}

Table~\ref{tab:per_question_overall_performance} shows that image-level captions consistently achieved higher human evaluation scores than region-level captions across most evaluation questions. The strongest performance for image-level captions was observed in scene context (Q7), hallucination check (Q1), color (Q4), and additional information (Q8), all exceeding 90\%. Region-level captions also performed well on color descriptions (Q4), but showed noticeably lower scores for additional information (Q8), actions (Q5), and object presence consistency (Q2). Overall, image-level captions achieved an average score of 78.96\%, compared to 57.17\% for region-level captions.

To demonstrate inter-annotator agreement, we used Fleiss' Kappa ($\kappa$) \cite{fleiss1971measuring} to measure agreement among the three annotators across the evaluation questions and caption types. Table~5 presents the corresponding agreement scores. To simplify interpretation, we adopted the agreement scale proposed by Landis and Koch \cite{landis1977measurement}: \textit{Almost Perfect ($>0.80$), Substantial (0.61--0.80), Moderate (0.41--0.60), Fair (0.21--0.40), Slight (0.01--0.20), and Poor Agreement ($\leq 0.00$)}.

\begin{table}[t]
\centering
\caption{Inter-Annotator Agreement (Fleiss' Kappa). Here, IC and RC denote image-level and region-level captions. Agreement strengths are denoted as follows: SL (Slight), FR (Fair), \underline{MD (Moderate)}, and \textbf{SB (Substantial)}.}
\label{tab:fleiss_kappa}

\footnotesize
\renewcommand{\arraystretch}{1.15}
\setlength{\tabcolsep}{3pt}

\begin{tabular*}{\columnwidth}{@{\extracolsep{\fill}}lcc}
\hline
\textbf{Evaluation Question} & \textbf{IC Kappa} & \textbf{RC Kappa} \\
\hline

Q1 (Hallucination Check) & 0.1947 (SL) & 0.3206 (FR) \\
Q2 (Object Presence Consist.) & 0.3394 (FR) & \underline{0.4929 (MD)} \\
Q3 (Size) & \underline{0.4856 (MD)} & \underline{0.5547 (MD)} \\
Q4 (Color) & \textbf{0.6302 (SB)} & \textbf{0.7051 (SB)} \\
Q5 (Action) & \textbf{0.6958 (SB)} & \underline{0.5175 (MD)} \\
Q6 (State) & 0.3811 (F) & 0.2388 (FR) \\
Q7 (Scene Context) & \textbf{0.7898 (SB)} & \textbf{0.6542 (SB)} \\
Q8 (Additional Information) & 0.3518 (F) & \underline{0.5197 (MD)} \\
\hline

\textbf{Overall Avg.} & \textbf{0.4836 (MD)} & \textbf{0.5004 (MD)} \\
\hline
\label{tab:inter_annotator_agreement}
\end{tabular*}
\end{table}

Based on the results in Table~\ref{tab:inter_annotator_agreement}, we observed three main findings. First, annotators showed the highest agreement when evaluating clear visual attributes such as color (Q4), actions (Q5), and scene/environment context (Q7) for both global and region-level captions. Second, annotators achieved consistent and reliable agreement when assessing object size (Q3). Third, the lowest agreement was observed for Q1, which requires verifying whether every mentioned element is present in the image, and Q6, which involves judging the exact state or condition of an object.

For global captions, annotators showed particularly strong agreement on scene context (Q7) and actions (Q5), but lower agreement on object presence verification (Q1). In contrast, region-level captions produced more balanced and stable agreement overall. Because annotators focused on specific image regions, they found it easier to verify objects (Q2) and evaluate additional contextual information (Q8).

\section{Experiments}

\subsection{Image Captioning Experiments}

%%%%%%%%%%%%%%%%%%%%%%%%%%%%%%%%%%%%%%%%%%%%%%%%%%%%%%%%%%%%%%%%%%%%%%%%%%%%%%%%%%%%%%%%%%%
% Switched to table* to span two columns
\begin{table*}[t]
\centering
\small
\setlength{\tabcolsep}{12pt} % Slightly increased spacing since it now has the full page width
% Force the table to snap to the full text width of the page
\begin{tabularx}{\textwidth}{>{\raggedright\arraybackslash}X c c c c c}
\hline
\textbf{Model} & \textbf{BLEU-4} & \textbf{METEOR} & \textbf{CIDEr-D} & \textbf{BERTScore} & \textbf{CLIPScore} \\
\hline
SmolVLM-256M-Instruct (Baseline)   & 0.05 & 0.30 & 0.05 & 0.86 & \underline{0.29} \\
SmolVLM-256M-Instruct (Finetuned)     & 0.26 & \textbf{0.50} & \underline{0.33} & \textbf{0.88} & \underline{0.29} \\
BLIP (Baseline)                    & 0.07 & 0.23 & 0.13 & 0.85 & 0.28 \\
BLIP (Finetuned)                      & 0.17 & 0.41 & 0.10 & 0.86 & 0.28 \\
BLIP2-2.2B (Baseline)              & 0.09 & 0.25 & 0.16 & 0.85 & \textbf{0.30} \\
BLIP2-2.2B (Finetuned)               & \underline{0.27} & 0.48 & \underline{0.33} & \underline{0.87} & 0.28 \\
Qwen2.5-VL-3B-Instruct (Baseline)   & \textbf{0.29} & \underline{0.49} & \textbf{0.36} & \textbf{0.88} & 0.28 \\
Qwen2.5-VL-3B-Instruct (Finetuned)    & 0.13 & 0.46 & 0.09 & \textbf{0.88} & \underline{0.29} \\
\hline
\end{tabularx}
\caption{Quantitative performance comparison of lightweight vision–language models on image-level captioning.}
\label{tab:vlm_captioning_results}
\end{table*}
%%%%%%%%%%%%%%%%%%%%%%%%%%%%%%%%%%%%%%%%%%%%%%%%%%%%%%%%%%%%%%%%%%%%%%%%%%%%%%%%%%%%%%%%%%%
As shown in Table~\ref{tab:vlm_captioning_results}, the pretrained variant of Qwen2.5-VL-3B-Instruct unexpectedly outperforms its finetuned counterpart across several captioning metrics. A likely reason is that the prompting strategy does not constrain caption length, the pretrained model retains its stronger general-purpose language generation capabilities, producing longer and more descriptive captions with greater lexical diversity and semantic coverage. Since metrics such as BLEU-4, METEOR, CIDEr-D, and BERTScore favor lexical overlap and detailed descriptions, the pretrained model achieves higher scores. In contrast, the finetuned model produces shorter and more concise captions after adapting to the training distribution, resulting in lower overlap-based metric scores despite maintaining comparable semantic alignment, as indicated by the similar BERTScore and CLIPScore values.

\begin{figure}[h]
\centering
\includegraphics[width=\columnwidth]{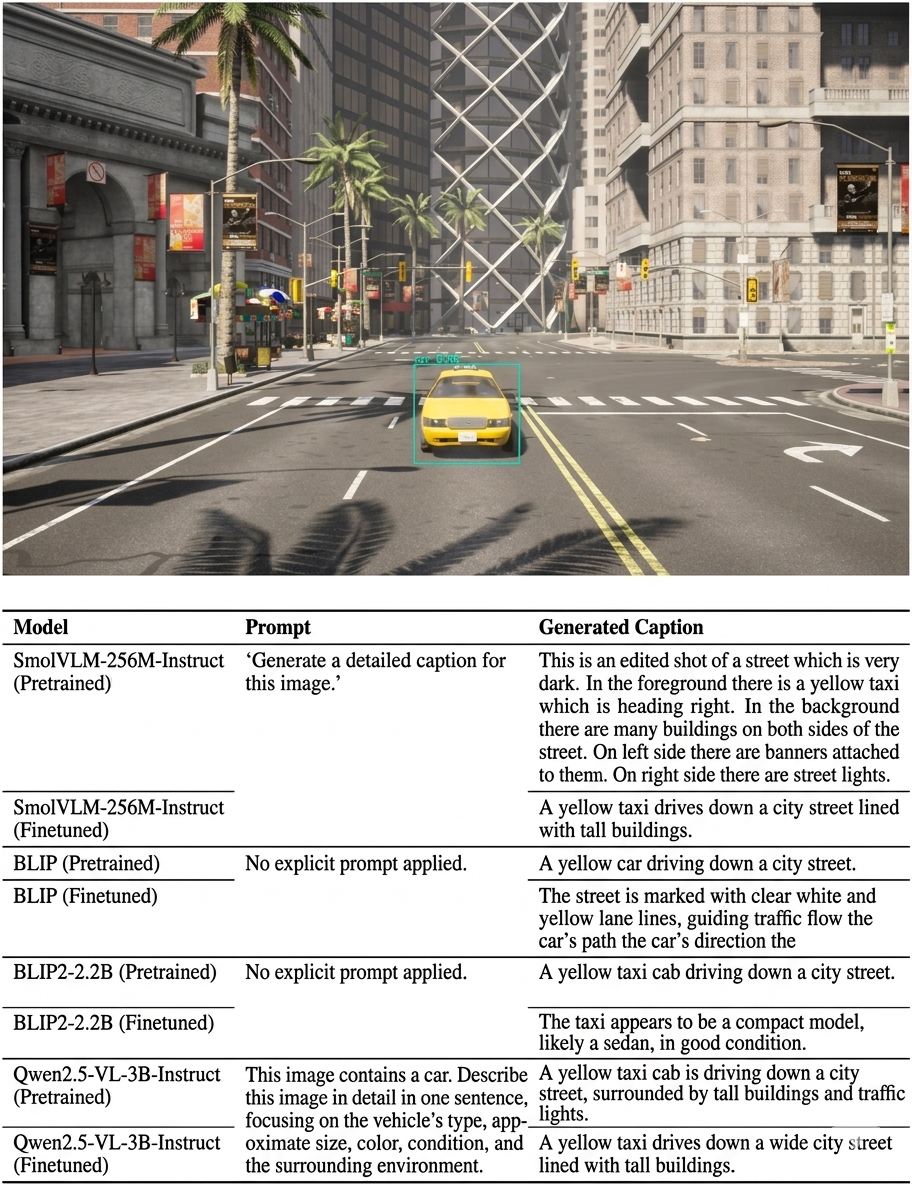}
\caption{Sample Image-Level Captions Generated by Pretrained and Fine-tuned Lightweight VLMs.}
\label{fig:experiments_sample_global_captions}
\end{figure}

Figure~\ref{fig:experiments_sample_global_captions} highlights several performance differences between pretrained and fine-tuned lightweight VLMs. The pretrained instruction-tuned models, particularly SmolVLM-256M-Instruct and Qwen2.5-VL-3B-Instruct, generated longer and more descriptive captions due to the use of explicit prompts requesting detailed scene understanding. These captions included richer contextual information such as \textit{buildings}, \textit{banners}, \textit{traffic lights}, and \textit{street layout}. In contrast, the corresponding fine-tuned models produced shorter and more object-centered captions, suggesting that fine-tuning encouraged stronger alignment toward concise dataset-style caption generation allowing the models to be able to describe fine-grained details including the \textbf{primary object} (\textit{taxi, taxi cab}), \textbf{action} (\textit{driving down, heading right}), \textbf{color} (\textit{yellow}), \textbf{size} and \textbf{environment} (\textit{street city with tall building}).
For BLIP and BLIP2, the pretrained models generated fluent but generic captions, the fine-tuned versions attempted to produce more fine-grained descriptions such as \textit{lane markings}, \textit{vehicle type}, and \textit{object condition}. However, the fine-tuned BLIP output became grammatically unstable, and the BLIP2 caption appeared incomplete, likely due to generation truncation from the \textit{max\_new\_tokens} limit which was set to 50. The bounding box in Figure~\ref{fig:experiments_sample_global_captions} is for illustrative purposes only. The actual raw image provided as input to the VLMs does not contain this visual marker.

\begin{figure}[h]
\centering
\includegraphics[width=\columnwidth]{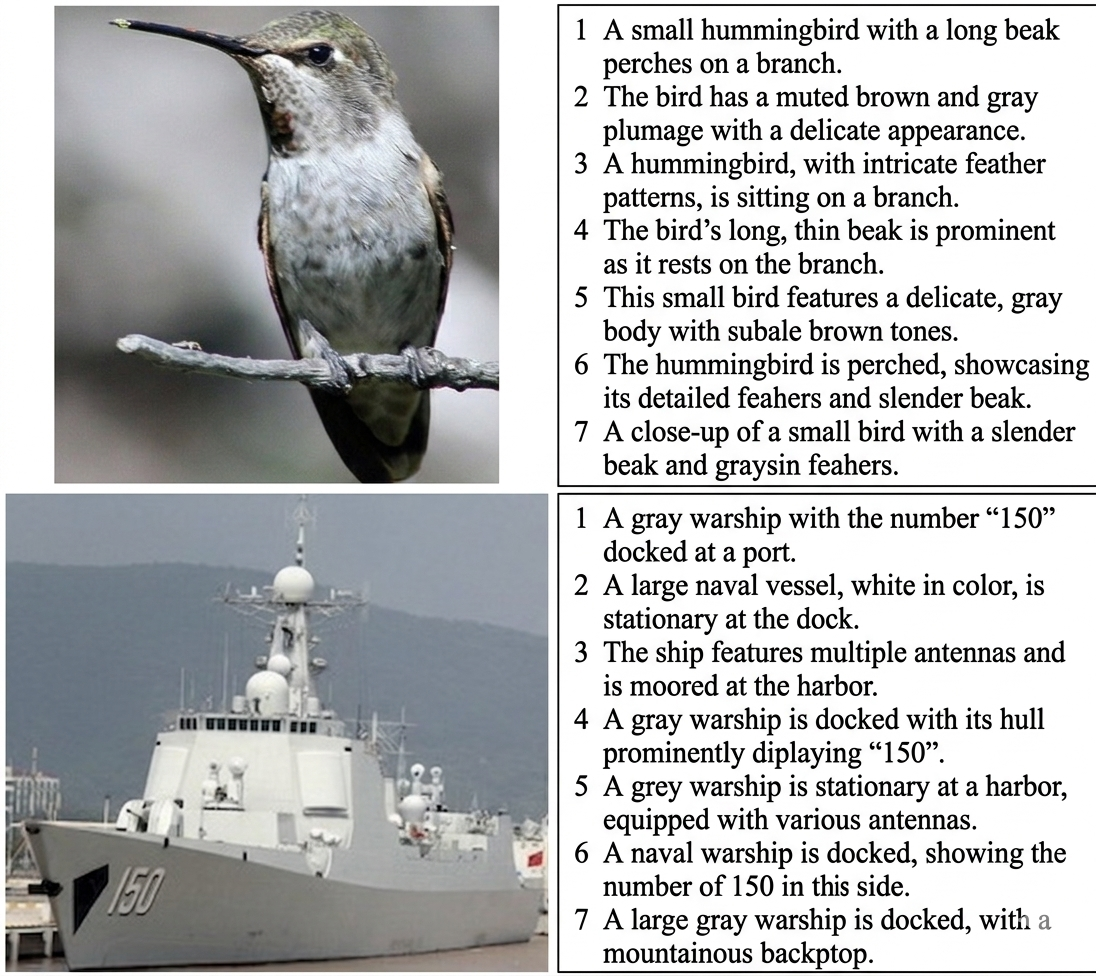}
\caption{Sample Region-Level Captions from MIRCaps.}
\label{fig:experiments_sample_region_captions}
\end{figure}

%%%%%%%%%%%%%%%%%%%%%%%%%%%%%%%%%%%%%%%%%%%%%%%%%%%%%%%%%%%%%%%%
%%%%%%%%%%%%%%%%%%%%%%%%%%%%%%%%%%%%%%%%%%%%%%%%%%%%%%%%%%%%%%%%
%%%%%%%%%%%%%%%%%%%%%%%%%%%%%%%%%%%%%%%%%%%%%%%%%%%%%%%%%%%%%%%%
%%%%%%%%%%%%%%%%%%%%%%%%%%%%%%%%%%%%%%%%%%%%%%%%%%%%%%%%%%%%%%%%
%%%%%%%%%%%%%%%%%%%%%%%%%%%%%%%%%%%%%%%%%%%%%%%%%%%%%%%%%%%%%%%%
%%%%%%%%%%%%%%%%%%%%%%%%%%%%%%%%%%%%%%%%%%%%%%%%%%%%%%%%%%%%%%%%

% ... End of your previous subsection ...

\subsection{Image Detection Experiments}

We evaluate two object detection frameworks, YOLOv12s and RT-DETR-L, on the same dataset. The dataset is split into 80\% training (112,269 images), 10\% validation (14,034 images), and 10\% testing (14,034 images). Both models are initialized with pretrained weights (YOLOv12s.pt and rtdetr-l.pt) and fine-tuned under identical experimental conditions. As shown in Table~\ref{tab:detection_comparison}, YOLOv12s achieves superior overall detection performance in terms of mAP and recall across all evaluation settings. RT-DETR-L, although trained for fewer epochs, still demonstrates competitive performance, suggesting potential gains with longer training, consistent with Zhao et al~\cite{zhao2024rtdetr}. We provide object detection experimental details, including training hyperparameters and sample object detection results, in Appendix~\ref{sec:appendix_experiments}.

%%%%%%%%%%%%%%%%%%%%%%%%%%%%%%%%%%%%%%%%%%%%%%%%%%%%%%%%%%%%%%%%
\begin{table*}[t]
\centering
\caption{Quantitative performance comparison of image detection models across datasets. \textbf{\# Images} and \textbf{Epochs} denote the total dataset size (in images) and the number of training epochs, respectively. The top performance is highlighted in \textbf{bold}, while the second-best performance is \underline{underlined}.}
\label{tab:detection_comparison}
\scriptsize
\renewcommand{\arraystretch}{1.1}

% Using tabularx to ensure the table matches the page width exactly
\begin{tabularx}{\textwidth}{@{} ll cccccccc X @{}}
\toprule
\textbf{Dataset} & \textbf{Model} & \textbf{\# Images} & \textbf{Epochs} & \textbf{mAP} & \textbf{mAP@50} & \textbf{mAP@75} & \textbf{mAP@50:95} & \textbf{Precision} & \textbf{Recall} & \textbf{IoU} \\
\midrule

DFire & TFNet (15.4M) & 9,869 & 200 & -- & \underline{81.2} & -- & \underline{46.8} & \underline{81.6} & 74.8 & -- \\

CrowdHuman & De-Homo (34.6M) & 19.4K & -- & \textbf{93.6} & -- & -- & -- & -- & -- & -- \\

OD-VIRAT-Tiny & RT-DETR-50 (41M) & 599,996 & 50 & 47.8 & 54.7 & 45.9 & -- & -- & -- & -- \\

OD-VIRAT-Tiny & YOLOX-Small (8.9M) & 599,996 & 50 & 35.9 & 48.5 & 39.5 & -- & -- & -- & -- \\

CUB-200-2011 & BSD-Net (30.6M) & 11,788 & 100 & 45.7 & 44.5 & -- & 41.9 & 48.2 & -- & -- \\

MS COCO (2014) & DEIM-D-FINE-L (31M) & 5,000 & 50 & 54.7 & 72.4 & \underline{59.4} & -- & -- & -- & -- \\

McShips & AK-DSAM-YOLOv13 (938M) & 14,709 & 200 & \underline{87.7} & \textbf{93.8} & \textbf{73.3} & -- & \textbf{93.3} & \textbf{89.8} & -- \\

VACaps (ours) & YOLOv12s (9.3M) & 141,364 & 92 & -- & 68.9 & -- & \textbf{49.7} & 80.7 & \underline{63.3} & \textbf{68.4} \\

VACaps (ours) & RT-DETR-L (32.9M) & 141,364 & 43 & -- & 63.6 & -- & 43.6 & 75.7 & 60.6 & \underline{47.8} \\

\bottomrule
\end{tabularx}
\end{table*}
%%%%%%%%%%%%%%%%%%%%%%%%%%%%%%%%%%%%%%%%%%%%%%%%%%%%%%%%%%%%%%%%

\section{Conclusion}
Our work introduced MIRCaps, a large-scale mixed-domain multimodal dataset containing 141,364 images, 981,947 image-level captions, 1,742,264 region-level captions, and 1,391,779 bounding box annotations spanning both general-purpose and CCTV-domain imagery. Experimental results on lightweight Vision-Language Models, including SmolVLM-256M-Instruct, BLIP, BLIP2, and Qwen2.5-VL-3B-Instruct, demonstrated their ability to generate improved descriptions with stronger emphasis on fine-grained attributes such as object category, action, color, size, state, and scene context. These findings indicate that the provided image-level and region-level captions can serve as effective supervision for training Vision-Language Models.

\section*{Limitations}
Although MIRCaps has proven effective in helping Vision-Language Models (particularly the lightweight VLMs evaluated in Section 4.1) generate higher-quality captions with focus on fine-grained attributes such as object category, action, color, size, state, and scene context, this research work presents the following limitations: 

\begin{enumerate}
    \item \textbf{Limited Bounding Box Coverage:} Although our dataset defines 15 object categories, several of the original source datasets contain fewer annotated categories. Experimental results using YOLOv12 and RT-DETR-L reveal reduced object detection performance caused by incomplete bounding box annotations, where not all instances of the 15 object categories are labeled in the original datasets. Future work should address this limitation by ensuring exhaustive annotation of all object occurrences.

    \item \textbf{Limited Fine-Tuning for Region-Level Captioning:} While the region-level captions achieved high semantic similarity with the corresponding cropped images, reflected by an average CLIPScore of 28.62, and were evaluated by human annotators to have good overall quality (57.17\%), the Vision-Language Models used for image-level captioning were not fine-tuned on these object-centric captions due to limited computational resources. Consequently, further investigation is needed to evaluate and improve the effectiveness of these models for region-level captioning tasks.    
\end{enumerate}

\section*{Ethical Considerations}

\textbf{Informed Consent:}
Human evaluators were informed about the potentially tedious nature of the caption evaluation task and were given the option to opt out at any time, although none chose to do so. They were research volunteers and were neither recruited through crowdsourcing platforms nor financially compensated. Therefore, payment details are not applicable.

\textbf{Potential Risks:}
The surveillance images in MIRCaps dataset carries risks of misuse in privacy-invasive applications. Additionally, source dataset aggregation introduces potential biases in scene and object distributions, which could affect downstream model fairness and generalization. Finally, automatically generated captions may contain residual hallucinations, and incomplete bounding box annotations could degrade downstream object detection performance.
{
    \small
    \bibliographystyle{ieeenat_fullname}
    \bibliography{main}
}
%%%%%%%%%%%%%%%%%%%%%%%%%%%%%%%%%%%%%%%%%%%%%%%%%%%%%%%%%%%%%%%%%%%%%%%%%%%%%%%
% APPENDIX START
%%%%%%%%%%%%%%%%%%%%%%%%%%%%%%%%%%%%%%%%%%%%%%%%%%%%%%%%%%%%%%%%%%%%%%%%%%%%%%%
\clearpage 
\appendix
\onecolumn % Switches from twocolumn layout to clear full-width presentation

\centerline{\Large \bf Appendix}
\vspace{1em}

\section{Introduction}
\label{sec:appendix_qualitative_analysis}
\begin{figure*}[h]
\centering
\includegraphics[width=0.80\textwidth]{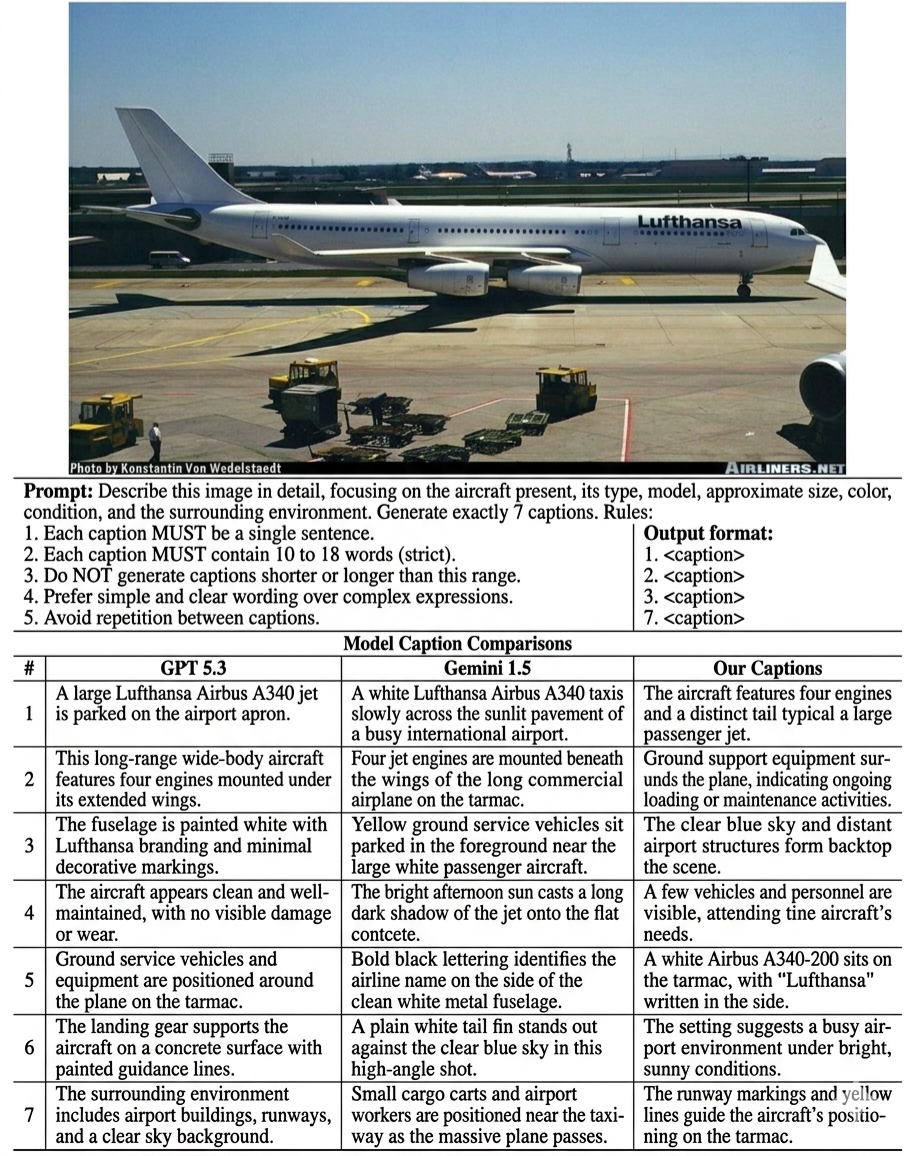}
\caption{Comparison between image-level captions generated by SOTA multimodal models and our curated descriptions.}
\label{fig:comparison_sota}
\end{figure*}

\noindent\textbf{Comparison of Image-Level Captions from SOTA LLMs and Our Dataset}.
Figure~\ref{fig:comparison_sota} provides a descriptive overview comparing image-level captions generated by state-of-the-art Large Language Models against our dataset descriptions.

\clearpage 
\section{Caption Generation}
\label{sec:appendix_caption_generation}
\begin{figure*}[h]
\centering
\includegraphics[width=0.9\textwidth]{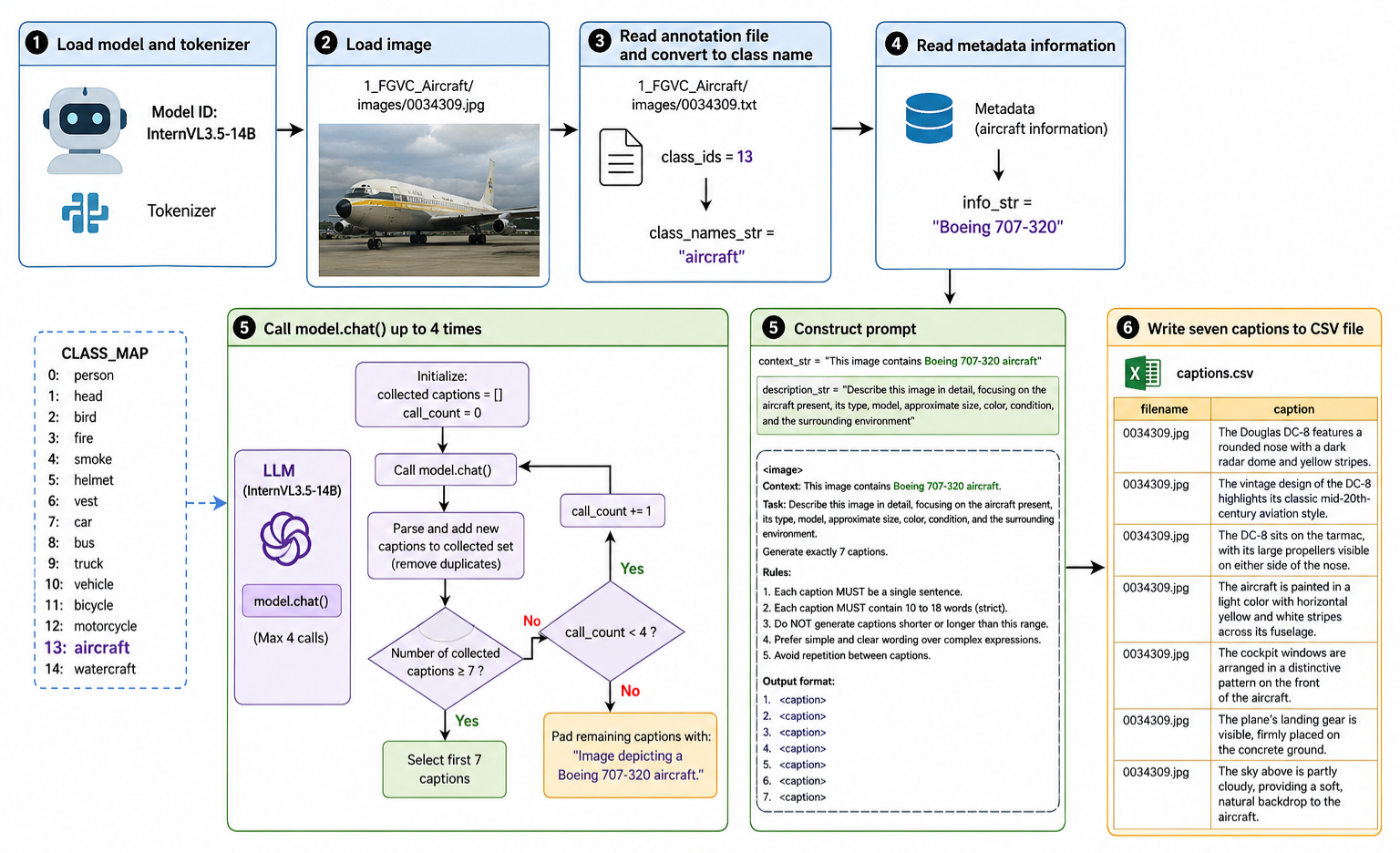} % Scales relative to the full text width
\caption{Caption Generation Pipeline.}
\label{fig:caption_generation_strategy_2}
\end{figure*}

\noindent\textbf{Caption Generation Strategy}. Figure~\ref{fig:caption_generation_strategy_2} illustrates the caption generation strategy applied for generation of both image and region-level captions as described in Section~\ref{sec:caption_generation_cleansing}. A sample image from FGVC-Aircraft dataset is used for illustrations purposes.

\clearpage
\section{Dataset Organization}
\label{sec:appendix_data_organization}
\begin{figure*}[h] 
\centering
\includegraphics[width=0.9\textwidth]{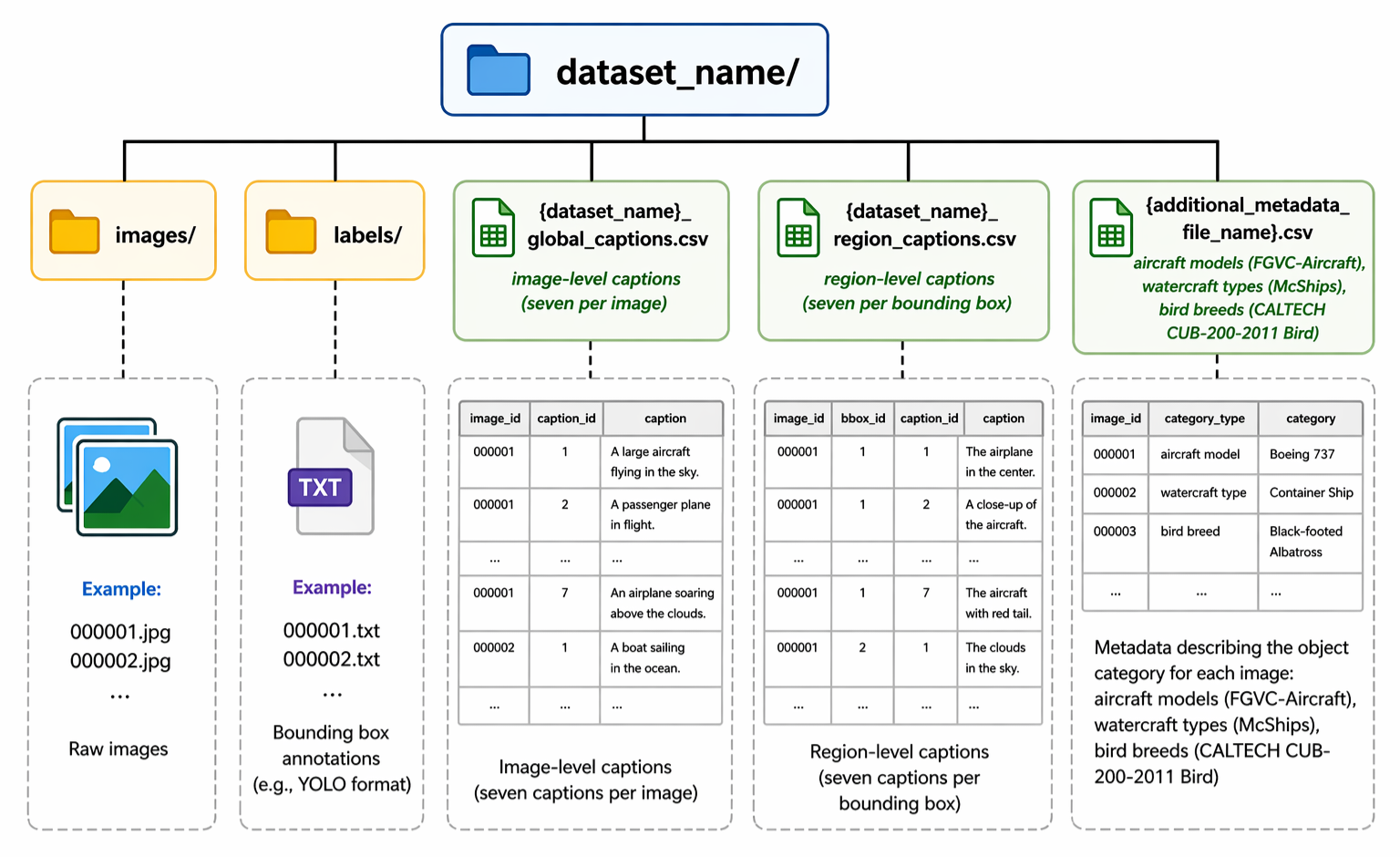} % Scales relative to the full page width
\caption{Basic Dataset Structure.}
\label{fig:appendix_dataset_structure}
\end{figure*}

\noindent\textbf{Dataset Folder Structure}. 
Figure~\ref{fig:appendix_dataset_structure} illustrates the dataset hierarchy discussed in Section~\ref{sec:dataset_organization_statistics}.

\clearpage
\section{Dataset Preprocessing}
\label{sec:appendix_dataset_preprocessing}
\noindent\textbf{Dataset Category ID Mapping.} Table~\ref{tab:category_id_mapping} details how original class labels and category IDs from various source datasets (\textbf{Before}) were consolidated and normalized into a single, unified label taxonomy (\textbf{After}) for effective downstream object detection tasks. For instance, datasets like FGVC-Aircraft and CALTECH CUB-200-2011, which originally lacked clear bounding box class indices (relying instead on mapped image IDs), are successfully brought into the bounding-box index system as \textbf{Aircraft (0)} and \textbf{Bird (2)}, respectively. Additionally, in the Vehicle Orientation dataset, class names such as \textit{Car\_front}, \textit{Car\_back}, and \textit{Car\_side} are merged under the unified label \textbf{Car (7)}.

\begin{table*}[h]
\centering
\small
\setlength{\tabcolsep}{6pt}
\renewcommand{\arraystretch}{1.1}
\begin{tabular}{p{5.0cm} p{5.5cm} p{5.5cm}}
\toprule
\textbf{Source Dataset} & \textbf{Before} & \textbf{After} \\
\midrule
FGVC-Aircraft dataset \cite{maji2013fgvc} & No specific class ID (bounding box mapped with image IDs) & Aircraft (0) \\ \midrule
DFire dataset \cite{venancio2022automatic} & Fire (0), Smoke (1) & Fire (3), Smoke (4) \\ \midrule
DFS Fire and Smoke dataset \cite{wu2022fire} & Fire (0), Smoke (1) & Fire (3), Smoke (4) \\ \midrule
CrowdHuman dataset \cite{shao2018crowdhuman} & fbox (full body), hbox (head) & Person (0), Head (1) \\ \midrule
OD-VIRAT-Tiny dataset \cite{ullah2011odvirat} & Bicycle (1), Car (2), Carrying\_object (3), Person (4), Vehicle (5) & Bicycle (11), Car (7), Person (0), Vehicle (10) \\ \midrule
Forest Fire dataset \cite{shamta2023forest} & Fire (0) & Fire (3) \\ \midrule
CQUniversity Fire and Smoke \cite{partheepan2025annotated} & Fire (0), Smoke (1) & Fire (3), Smoke (4) \\ \midrule
CALTECH CUB-200-2011 \cite{wah2011cub} & No specific class ID (bounding box mapped with image IDs) & Bird (2) \\ \midrule
Vehicle Orientation dataset \cite{kumar2022traffic} & Car\_front, Car\_back, Car\_side, Motorcycle\_front, Truck\_front, Bicycle\_front & Car (7), Motorcycle (12), Truck (9), Bicycle (11) \\ \midrule
MS COCO (2014) dataset \cite{lin2014coco} & Bicycle (2), Motorcycle (4), Bus (6), Truck (8) & Bicycle (11), Motorcycle (12), Bus (8), Truck (9) \\ \midrule
McShips dataset \cite{zheng2020mcships} & Civilianship, Warship & Watercraft (14) \\ \midrule
Personal Protective Equipment \cite{huang2025ppe} & Helmet (0), Vest (3) & Helmet (5), Vest (6) \\ \midrule
HardHat-Vest dataset \cite{zahit2023hardhat} & Helmet (0), Vest (1), Head (2) & Helmet (5), Vest (6), Head (1) \\
\bottomrule
\end{tabular}
\caption{Dataset class remapping from original annotations (Before) to unified label space (After) across multiple datasets.}
\label{tab:category_id_mapping}
\end{table*}

\section{Image Quality Assessment}
\label{sec:image_quality_assessment}

\subsection{Non-Reference Image Quality and Visual Attribute Analysis}
As shown in Table~\ref{tab:nriqa}, the dataset demonstrates generally good perceptual quality (NIQE = 5.75, BRISQUE = 21.08) with moderate visible degradation (PIQE = 41.86). 

\subsection{Visual Attribute Analysis}
The dataset maintains normal brightness and contrast levels and moderate-to-high sharpness, as presented in Table~\ref{tab:visual_attributes}, making it highly suitable for standard vision-language tasks.

\subsection{Dataset Image Characteristics}
As shown in Table~\ref{tab:image_dataset_characteristics}, the dataset contains images with highly diverse resolutions (ranging from 60$\times$60 to 7200$\times$10800, with a mean of 698$\times$1056), indicating substantial variability in size, while the average aspect ratio of 1.49 aligns closely with typical natural image datasets. Color statistics (mean RGB values around 109--118 with standard deviations $\sim$56--58) suggest well-balanced, natural image distributions without extreme lighting conditions. Additionally, the dataset shows strong data integrity, with no corrupted images detected, making it reliable and representative for real-world vision tasks. 

\begin{table}[h]
\centering
\small % Scaled up from \scriptsize for clean readability
\caption{Non-Reference Image Quality Assessment (NR-IQA)}
\label{tab:nriqa}
\begin{tabularx}{\textwidth}{l l X} % X automatically fills the rest of the page width
\hline
\textbf{Metric} & \textbf{Avg. Value} & \textbf{Description} \\
\hline
NIQE~\cite{mittal2013blind} & 5.75 & Best quality (natural images): 0--4 \\ 
& & \textbf{Good quality images: 4--6} \\ 
& & Moderate/degraded quality: 6--10 \\ 
& & Poor quality (noisy, blurred, synthetic artifacts): 10+ \\ \hline
BRISQUE~\cite{mittal2012noreference} & 21.08 & 0--100 scale. Lower values indicate better image quality. \\ 
& & Excellent/high-quality images: 0--20 \\ 
& & \textbf{Good quality images: 20--40} \\ 
& & Moderate quality (noticeable distortions): 40--60 \\ 
& & Poor quality images: 60--80 \\ 
& & Very poor/heavily distorted images: 80--100 \\ \hline
PIQE~\cite{venkatanath2015blind} & 41.86 & 0--100 scale. Lower values indicate better image quality. \\ 
& & High-quality images: 0--10 \\ 
& & Good quality images: 10--20 \\ 
& & Moderate quality (minor visible artifacts): 20--35 \\ 
& & \textbf{Low-quality images (noticeable degradation): 35--50} \\ 
& & Very poor-quality images: 50--70 \\ 
& & Severely distorted/unusable images: 70--100 \\
\hline
\end{tabularx}
\end{table}

\begin{table}[h]
\centering
\small % Scaled up from \scriptsize for better readability
\caption{Visual Attribute Analysis}
\label{tab:visual_attributes}
% Widths increased by another 0.25cm (from 1.75cm and 1.50cm)
\begin{tabularx}{\textwidth}{p{2.00cm} p{1.75cm} X} 
\hline
\textbf{Metric} & \textbf{Avg. Value} & \textbf{Description} \\
\hline
Brightness & 115.62 & Dark images (low brightness): 0--85 \\ 
& & \textbf{Normal brightness (most natural images): 85--170} \\ 
& & Bright images (high exposure): 170--255 \\ \hline
Contrast & 56.20 & Low contrast images: 0--30 \\ 
& & \textbf{Normal contrast images: 30--70} \\ 
& & High contrast images: 70--128+ \\ \hline
Sharpness & 1203.04 & Low sharpness (blurred): 0--400 \\ 
& & Moderate sharpness (natural images): 400--1200 \\ 
& & \textbf{High sharpness (high-quality images): 1200--3000+} \\
\hline
\end{tabularx}
\end{table}

\begin{table*}[t]
\centering
\small % Clean, highly readable font size
\caption{Image Dataset Characteristics.}
\label{tab:image_dataset_characteristics}
% First three columns automatically size to text width; final column dynamically fills the rest
\begin{tabular}{l l l p{\dimexpr\textwidth-2.3cm-1.8cm-4.5cm-6\tabcolsep\relax}}
\hline
\textbf{Category} & \textbf{Metric} & \textbf{Value} & \textbf{Description} \\
\hline

\multirow{6}{*}{Resolution}
& Min  & 60 $\times$ 60 & -- \\
& Max  & 7200 $\times$ 10800 & -- \\
& Mean & 698.08 $\times$ 1056.07 & -- \\
& Std. & 383.44 $\times$ 640.25 & Small variation: Width 0--100 px / Height 0--80 px \\
&      &                    & Medium variation: Width 300--800 px / Height 200--700 px \\
&      &                    & High variation: Width 800--2000+ px / Height 700--1500+ px \\
\hline

\multirow{3}{*}{Aspect Ratio}
& Mean & 1.49 & Curated/resized image dataset: 1.0--1.2 \\
&      &      & Natural image datasets: 1.2--1.6 (typical real datasets) \\
&      &      & Web-scale mixed datasets: 1.3--2.0+ \\
\hline

\multirow{2}{*}{Color (RGB)}
& Mean & R: 118.029, G: 115.61, B: 109.34
& Natural images: R 110--140 / G 105--135 / B 100--130 \\
& Std. & R: 58.33, G: 56.69, B: 58.15
& Natural images: R 40--60 / G 40--60 / B 40--65 \\
\hline

Data Integrity & Corrupted Rate & 0.0\% & No corrupted images detected. \\
\hline

\end{tabular}
\end{table*}

\clearpage
\subsection{Human Evaluation of Caption Quality}
\label{sec:appendix_human_evaluation_caption_quality}
\noindent\textbf{Human Evaluation of Image and Region-Level Captions.}
To verify image-caption semantic correspondence and especially their alignment with the eight dimensions mentioned in Section~\ref{sec:human_evaluation_caption_quality}
(\textit{Hallucination Check}, \textit{Object Presence Consistency}, \textit{Size}, \textit{Color}, \textit{Action}, \textit{State}, \textit{Scene Context}, and \textit{Additional Information}), we perform human evaluation across image-level captions (Table~\ref{tab:image_level_caption_performance}), and region-level captions (Table~\ref{tab:region_level_caption_performance}).

\begin{table}[h]
\centering
\caption{Performance of Human Evaluation of Image-Level Caption}
\label{tab:image_level_caption_performance}
\small
\renewcommand{\arraystretch}{1.15}
\setlength{\tabcolsep}{8pt}
\begin{tabular}{lcccccccc}
\hline
\textbf{ID} & \textbf{Q1} & \textbf{Q2} & \textbf{Q3} & \textbf{Q4} & \textbf{Q5} & \textbf{Q6} & \textbf{Q7} & \textbf{Q8} \\
\hline
1 & 0.94 & 0.95 & 0.11 & 0.92 & 0.60 & 0.62 & 0.99 & 0.92 \\
2 & 0.98 & 0.89 & 0.29 & 0.95 & 0.65 & 0.72 & 0.99 & 0.98 \\
3 & 1.00 & 0.93 & 0.26 & 0.95 & 0.59 & 0.80 & 0.99 & 0.96 \\
\hline
\textbf{Avg.} & \textbf{0.97} & \textbf{0.92} & \textbf{0.22} & \textbf{0.94} & \textbf{0.61} & \textbf{0.71} & \textbf{0.99} & \textbf{0.95} \\
\hline
\end{tabular}
\end{table}

\begin{table}[h]
\centering
\caption{Performance of Human Evaluation of Region-Level Caption}
\label{tab:region_level_caption_performance}
\small
\renewcommand{\arraystretch}{1.15}
\setlength{\tabcolsep}{8pt}
\begin{tabular}{lcccccccc}
\hline
\textbf{ID} & \textbf{Q1} & \textbf{Q2} & \textbf{Q3} & \textbf{Q4} & \textbf{Q5} & \textbf{Q6} & \textbf{Q7} & \textbf{Q8} \\
\hline
1 & 0.55 & 0.71 & 0.23 & 0.88 & 0.25 & 0.31 & 0.55 & 0.19 \\
2 & 0.74 & 0.69 & 0.45 & 0.93 & 0.43 & 0.67 & 0.61 & 0.39 \\
3 & 0.94 & 0.75 & 0.38 & 0.92 & 0.45 & 0.74 & 0.58 & 0.40 \\
\hline
\textbf{Avg.} & \textbf{0.74} & \textbf{0.72} & \textbf{0.35} & \textbf{0.91} & \textbf{0.37} & \textbf{0.57} & \textbf{0.58} & \textbf{0.33} \\
\hline
\end{tabular}
\end{table}

\clearpage
\section{Object Detection Experimental Details}
\label{sec:appendix_experiments}

\noindent\textbf{YOLOv12s Configuration.}
We use stochastic gradient descent (SGD) with an initial learning rate of 0.01 and cosine decay scheduling. The model is trained for 92 epochs with a batch size of 12. Early stopping with a patience boundary of 50 epochs is applied. Mosaic data augmentation routines are explicitly disabled during the final 10 training epochs to stabilize model weight convergence.
\medskip % Adds a clean, medium-sized vertical space before the next block

\noindent\textbf{RT-DETR-L Configuration.}
We use AdamW with an initial learning rate of 1e-4 and cosine learning rate scheduling with warmup, training for 43 epochs with a batch size of 2 due to hardware constraints. 
\medskip % Adds a clean, medium-sized vertical space before the next block

\noindent\textbf{Common Configuration.}
The input resolution is fixed at 640$\times$640 for all baseline experiments. Mixed precision training (AMP) is enabled to improve processing efficiency. 

\begin{figure}[h]
\centering
\includegraphics[width=0.95\columnwidth]{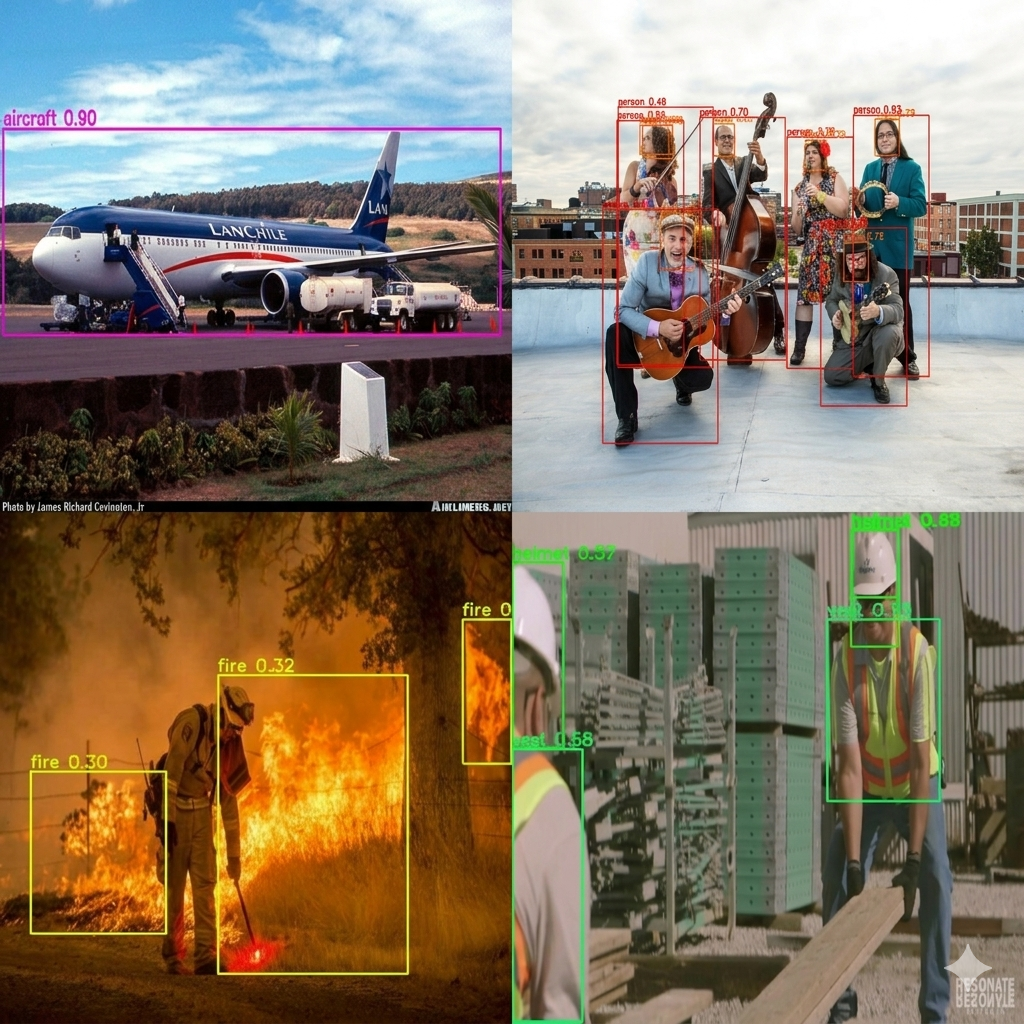}
\caption{Sample Image Detection Results from YOLOv12s.}
\label{fig:experiments_image_detection}
\end{figure}

\clearpage

\section{Dataset Sources, Licensing, and Download URLs}
\label{sec:appendix_dataset_sources}

Our dataset is released under the Creative Commons Attribution 4.0 International (CC BY 4.0) license. However, it is not redistributed in its full form due to licensing restrictions associated with several constituent datasets, including FGVC-Aircraft (non-commercial research-only license), DFS Fire and Smoke (No explicit license specified), CrowdHuman (non-commercial research and educational use only license), OD-VIRAT-Tiny (non-commercial research-only license), and McShips (No explicit license specified). For these source datasets, we provide only the caption annotations and the available metadata files (FGVC-Aircraft, McShips). Comprehensive scripts are detailed to assist users in fetching corresponding images and bounding annotations directly from official hosts. 

All primary assets remain subject to their original licenses. Users are required to adhere to the distribution restrictions outlined in the table below (Table~\ref{tab:dataset_sources}).

\begin{table*}[h]
\centering
\scriptsize
\caption{Dataset Sources, Licensing, and Download URLs. License entries shown in bold indicate datasets explicitly restricted from third-party distribution.}
\label{tab:dataset_sources}

\begin{tabular}{p{3.5cm} p{2.8cm} p{5.0cm} p{4.5cm}}
\hline
\textbf{Dataset} & \textbf{License} & \textbf{Download URL} & \textbf{Instruction} \\
\hline

FGVC-Aircraft dataset \cite{maji2013fgvc}
& \textbf{Non-commercial research-only}
& \url{https://www.robots.ox.ac.uk/~vgg/data/fgvc-aircraft}
& Select ``Data, annotations, and evaluation code'' to download. \\
\hline

DFire dataset \cite{venancio2022automatic}
& CC0 1.0 Universal
& \url{https://github.com/gaia-solutions-on-demand/DFireDataset}
& -- \\
\hline

DFS Fire and Smoke \cite{wu2022fire}
& \textbf{No explicit license}
& \url{https://universe.roboflow.com/cv-atyqk/dfs-fire/dataset/3}
& Sourced via Roboflow repository path. \\
\hline

CrowdHuman dataset \cite{shao2018crowdhuman}
& \textbf{Non-commercial research/edu}
& \url{https://www.kaggle.com/datasets/leducnhuan/crowdhuman}
& Download training partitions. \\
\hline

OD-VIRAT-Tiny \cite{ullah2011odvirat}
& \textbf{Non-commercial research-only}
& \url{https://drive.google.com/drive/folders/1MqVKIfS_RimUVVin1UHk_uwPmex5vid7}
& Sourced via original Google Drive partition. \\
\hline

Forest Fire dataset \cite{shamta2023forest}
& CC BY 4.0
& \url{https://data.mendeley.com/datasets/fcsjwd9gr6}
& Extract \textit{ForesFireDataset(ObjectDetection).zip}. \\
\hline

CQUniversity Fire/Smoke \cite{partheepan2025annotated}
& CC BY 4.0
& \url{https://acquire.cqu.edu.au/articles/dataset/Annotated_Fire_-Smoke_Image_Dataset_for_fire_detection_Using_YOLO_/28747046}
& Extract \textit{MyData\_Fire.zip}. \\
\hline

CALTECH CUB-200-2011 \cite{wah2011cub}
& CC BY
& \url{https://data.caltech.edu/records/65de6-vp158}
& Fetch \textit{CUB\_200\_2011.tgz}. \\
\hline

Vehicle Orientation \cite{kumar2022traffic}
& MIT License
& \url{https://github.com/sekilab/VehicleOrientationDataset}
& Pull down \textit{training\_2.zip}. \\
\hline

MS COCO (2014) \cite{lin2014coco}
& CC-BY 4.0
& \url{http://images.cocodataset.org/zips/train2014.zip}
& Training and validation targets. \\
\hline

McShips dataset \cite{zheng2020mcships}
& \textbf{No explicit license}
& \url{https://github.com/ZhengYitong2333/Mcships}
& Sourced via citation mapping guidelines. \\
\hline

PPE Dataset \cite{huang2025ppe}
& CC BY 4.0
& \url{https://data.mendeley.com/datasets/zkzghjvpn2/6}
& Extract \textit{20250731-PPE2286y.zip}. \\
\hline

HardHat-Vest Dataset \cite{zahit2023hardhat}
& CC0: Public Domain
& \url{https://www.kaggle.com/datasets/muhammetzahitaydn/hardhat-vest-dataset-v3}
& Kaggle distribution platform link. \\
\hline

\end{tabular}
\end{table*}

\end{document}